\documentclass[sigconf]{acmart}

\usepackage{subfigure}
\usepackage[export]{adjustbox}
\usepackage{enumitem}
\usepackage{bbm}
\usepackage{multirow}
\usepackage{makecell}

\AtBeginDocument{%
  \providecommand\BibTeX{{%
    \normalfont B\kern-0.5em{\scshape i\kern-0.25em b}\kern-0.8em\TeX}}}

\setcopyright{acmlicensed}
\copyrightyear{2023}
\acmYear{2023}

\begin{document}

\title{Scene-Generalizable Interactive Segmentation of Radiance Fields}





\author{Songlin Tang}
\affiliation{
  \institution{Harbin Institute of Technology \city{Shenzhen} \country{China}}
}
\email{21s15092@stu.hit.edu.cn}

\author{Wenjie Pei}
\authornote{Corresponding author.}
\affiliation{
  \institution{Harbin Institute of Technology \city{Shenzhen} \country{China}}
}
\email{wenjiecoder@outlook.com}

\author{Xin Tao}
\affiliation{%
 \institution{Y-tech, Kuaishou Technology}
 \city{Beijing} \country{China}
 } 
\email{jiangsutx@gmail.com}

\author{Tanghui Jia}
\affiliation{
  \institution{Harbin Institute of Technology \city{Shenzhen} \country{China}}
}
\email{200111413@stu.hit.edu.cn}

\author{Guangming Lu}
\affiliation{
  \institution{Guangdong Provincial Key Laboratory of Novel Security Intelligence Technologies}
  \institution{Harbin Institute of Technology \city{Shenzhen} \country{China}}
}
\email{luguangm@hit.edu.cn}

\author{Yu-Wing Tai}
\affiliation{%
 \institution{Dartmouth College}
 \city{Hanover} \country{United States}
 } 
\email{yuwing@gmail.com}


\begin{abstract}
  Existing methods for interactive segmentation in radiance fields entail scene-specific optimization and thus cannot generalize across different scenes, which greatly limits their applicability. In this work we make the first attempt at Scene-Generalizable Interactive Segmentation in Radiance Fields (\emph{SGISRF}) and propose a novel \emph{SGISRF} method, which can perform 3D object segmentation for novel (unseen) scenes represented by radiance fields, guided by only a few interactive user clicks in a given set of multi-view 2D images. In particular, the proposed \emph{SGISRF} focuses on addressing three crucial challenges with three  specially designed techniques. First, we devise the Cross-Dimension Guidance Propagation to encode the scarce 2D user clicks into informative 3D guidance representations. Second, the Uncertainty-Eliminated 3D Segmentation module is designed to 
achieve efficient yet effective 3D segmentation. Third, Concealment-Revealed Supervised Learning scheme is proposed to reveal and correct the concealed 3D segmentation errors resulted from the supervision in 2D space with only 2D mask annotations.
Extensive experiments on two real-world challenging benchmarks covering diverse scenes demonstrate 1) effectiveness and scene-generalizability of the proposed method, 2) favorable performance compared to classical method requiring scene-specific optimization.
  \vspace{-12pt}
\end{abstract}


\vspace{-15pt}
\keywords{Interactive segmentation, Radiance fields, Scene-generalizable}


\begin{teaserfigure}
\vspace{-10pt}
  \centering
  \includegraphics[width=0.95\textwidth]{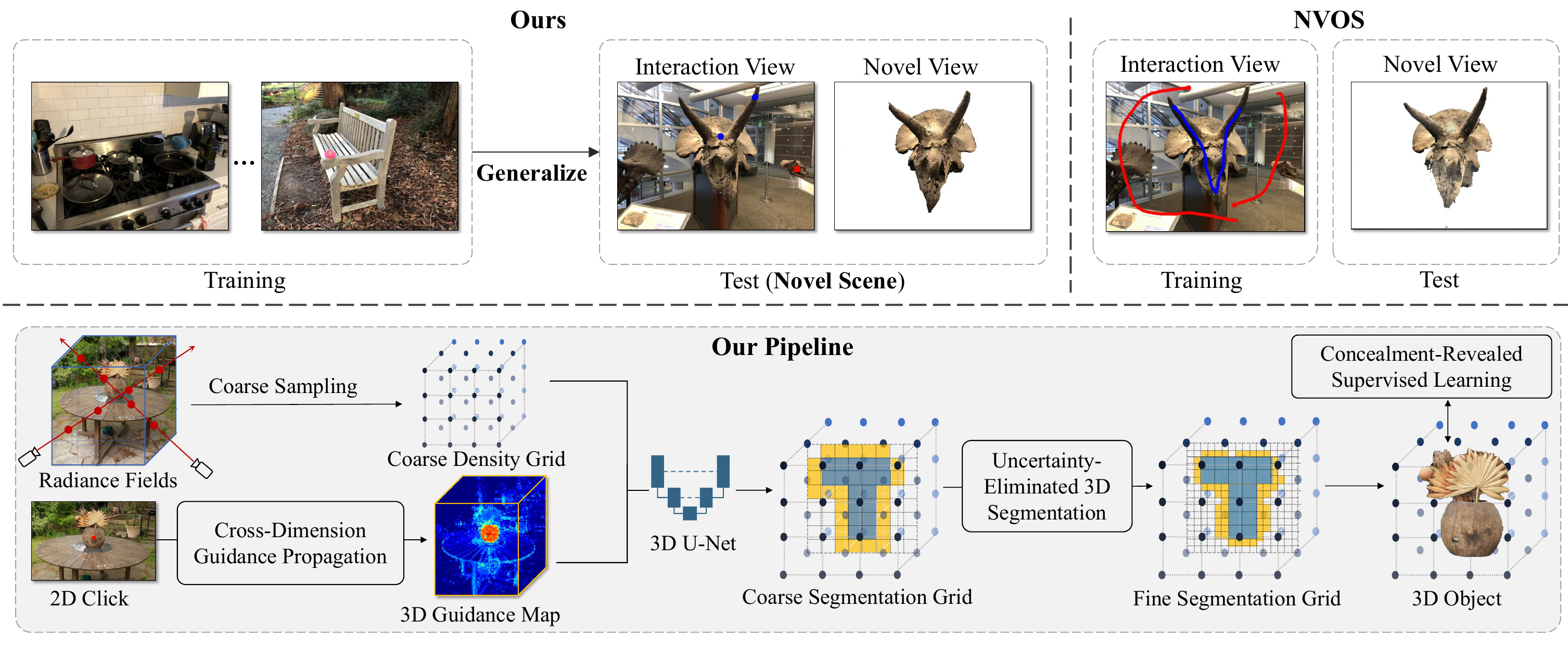}
  \vspace{-12pt}
  \caption{Our \emph{SGISRF} is able to generalize across different scenes and performs segmentation on novel scenes whilst NVOS demands scene-specific optimization. Nevertheless, our method still outperforms NVOS, even with less user guidance (only a few user clicks rather than long user strokes).}
  \label{fig:teaser}
\end{teaserfigure}

 \settopmatter{printacmref=false} 
\renewcommand\footnotetextcopyrightpermission[1]{}

\maketitle

\section{Introduction}
Interactive segmentation of radiance fields aims to perform 3D scene segmentation in radiance fields, guided by user interactive annotations such as point clicks and strokes in a few views of 2D images. It allows users to perform 3D editing of radiance fields or 2D view-consistent image editing like object selection, which is particularly useful in scene understanding or manipulation. 2D interactive segmentation has achieved considerable progress~\cite{sofiiuk2022reviving, lin2022focuscut, liu2022pseudoclick} whilst 3D interactive segmentation in radiance fields is a newly emerging task that remains a research challenge. 

Existing methods for interactive segmentation of radiance fields entail scene-specific optimization and thus cannot generalize across different scenes, which greatly limits their applicability. A prominent example is NVOS~\cite{ren2022neural}, the first method for this task, which projects the 2D user scribbles into 3D voxels and utilizes them as to learn a binary classifier between foreground and background for 3D segmentation. Formulating segmentation as a binary classification task in such a way, NVOS cannot be applied to novel scenes. Besides, learning an effective classifier requires a large number of training samples which correspond to the same amount of user interactions. Furthermore, NVOS demands 3D graph-cut for post-processing, which limits its efficiency and practicability. Another state-of-the-art method for interactive segmentation of radiance field is ISRF~\cite{goel2022interactive}, which seeks to learn 3D semantic features by distillation from pre-learned 2D semantic features. As a result, ISRF also relies on scene-specific optimization in that such feature distillation has to be conducted for each scene individually. To circumvent the limitation of demanding scene-specific optimization,
we develop a novel methods for scene-generalizable interactive segmentation in radiance fields (\emph{SGISRF}), which is able to perform interactive 3D segmentation for unseen scenes, as illustrated in Figure~\ref{fig:teaser}.

Effective modeling of the proposed \emph{SGISRF} requires addressing three crucial challenges. 1) How to encode the scarce 2D user clicks into informative 3D guidance features so as to achieve precise 3D segmentation with minimum user guidance? 2) Performing segmentation in higher-resolution 3D grid space yields more precise performance whilst incuring heavier computational burden,
which poses the demand of efficient yet effective 3D segmentation algorithms. 3) Since only 2D mask annotations are available for optimization of our \emph{SGISRF}, the predicted 3D segmentation results need to be rendered into 2D space for supervision. However, substantial 3D segmentation details are lost during such rendering, which may conceal some prediction errors of 3D segmentation and fail to correct these error during optimization.

We propose three novel technical designs to tackle above three challenges correspondingly. First of all, we devise the Cross-Dimension Guidance Propagation to encode the user clicks. It first propagates the scattered user click points in 2D space by leveraging the semantic similarities, then lifts the propagated guidance from 2D to 3D space and further performs 3D propagation by exploiting the geometric continuities. 
Secondly, to perform 3D segmentation efficiently yet accurately, we particularly design an Uncertainty-Eliminated 3D Segmentation module based on 3D U-Net~\cite{cciccek20163d}, which first predicts a coarse mask in a low-resolution 3D feature space and then identifies and refines a fraction of uncertain mask area in high-resolution 3D feature space. 
Thirdly, we propose the Concealment-Revealed Supervised Learning scheme to reveal and correct the concealed 3D segmentation errors resulted from the supervision in 2D space with only 2D mask annotations. To conclude, the main contributions of this work are summarized as follows:
\begin{itemize}[leftmargin =*, itemsep = 0pt, topsep = -2pt]
    \item To the best of our knowledge, our \emph{SGISRF} is the first proposed method for interactive segmentation of radiance fields that is able to generalize to novel scenes.
    \item Benefiting from the designed Cross-Dimension Guidance Propagation, \emph{SGISRF} is able to perform 3D segmentation with only a few user clicks, which enables much more efficient user interaction than other state-of-the-art methods such as NVOS and ISRF requiring a number of user strokes. 
    \item Owing to the proposed Uncertainty-Eliminated 3D segmentation framework, our \emph{SGISRF} can perform 3D segmentation quite efficiently with no post-processing required.
    \item We design Concealment-Revealed Supervised Learning scheme, which enables effective supervision with 2D mask annotations.
    \item Extensive experiments on two challenging datasets covering diverse scenes demonstrate the effectiveness and well scene-generalizability of the proposed \emph{SGISRF}. Notably, the generalization performance of our \emph{SGISRF} on unseen scenes even surpasses substantially the performance of NVOS which performs scene-specific optimization on the same set of test scenes. 
\end{itemize}

\section{Related Work}
\noindent\textbf{Neural Radiance Fields} (NeRF)~\cite{mildenhall2021nerf} was proposed for novel view synthesis by optimizing a differentiable volume rendering. Due to excellent ability of 3D scene representation and high-quality rendering of images, NeRF has made a remarkable progress in the past few years. In particular, many variants of NeRF~\cite{muller2022instant, fridovich2022plenoxels, sun2022direct, wizadwongsa2021nex, xu2022point} has been developed to deal with different technical challenges~\cite{tancik2022block, martin2021nerf, mildenhall2022nerf, wang2021neus, chen2021mvsnerf} or be applied to different tasks~\cite{hu2022nerf, kundu2022panoptic, poole2022dreamfusion, jambon2023nerfshop, kerr2023lerf}. A key drawback of NeRF lies in the slow rendering speed. Explicit representation~\cite{fridovich2022plenoxels} and Hybrid representations~\cite{sun2022direct, muller2022instant, chen2022tensorf} have been proposed to speed up rendering while maintaining high storage efficiency. We adopt TensoRF ~\cite{chen2022tensorf} as our radiance fields for scene representation due to its fast rendering speed and efficient storage.

\begin{figure*}[!t]
  \centering
  \includegraphics[width=0.98\textwidth]{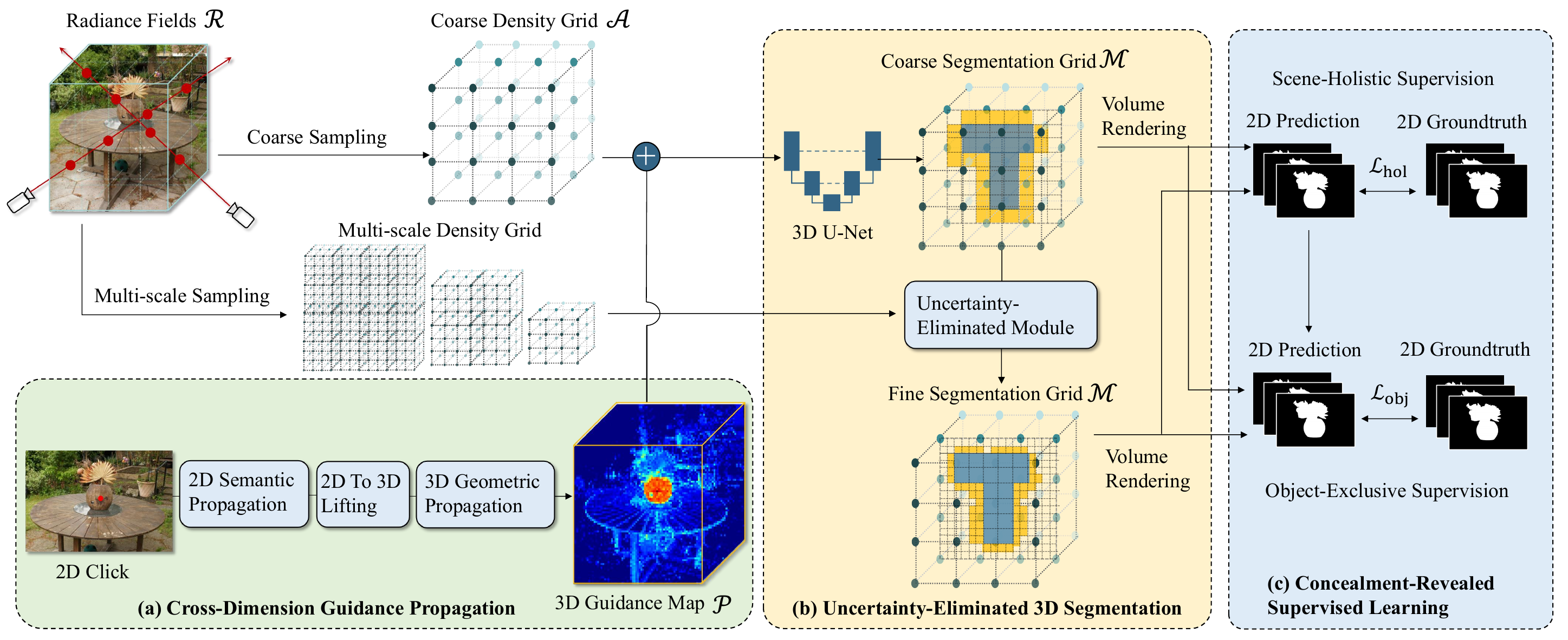}
  \vspace{-13pt}
  \caption{Overview of the proposed \emph{SGISRF}. It encodes the user clicks in 2D space into informative 3D guidance features by Cross-Dimension Guidance Propagation, and fuses them with the density features extracted from the radiance fields. The fused features are fed into the Uncertainty-Eliminated 3D segmentation module for efficient yet accurate segmentation. The whole model is optimized effectively with 2D mask annotations under the designed Concealment-Revealed Supervised Learning.}
  \vspace{-9pt}
  \label{fig:pineline}
\end{figure*}

\noindent\textbf{Interactive 2D Image Segmentation} is a classical vision task which has been well studied. Early traditional methods, e.g., Graph Cut~\cite{boykov2001interactive} and GrabCut~\cite{rother2004grabcut}, require to carefully design the energy optimization function and minimum-cut optimization algorithm. Recent deep learning based methods~\cite{xu2016deep, jang2019interactive, lin2020interactive, zhang2020interactive, chen2021conditional, sofiiuk2022reviving} rely on a large amount of annotated data to learn a generalizable model and can achieve more accurate segmentation than traditional methods. They employ existing mature deep learning networks for semantic segmentation or instance segmentation as backbone, and encode user interactions (like clicks or strokes) into additional guidance features to guide the learning of models iteratively. 

\noindent\textbf{Interactive Radiance Fields Segmentation} is a newly emerging task that remains a research challenge. NVOS~\cite{ren2022neural} makes the first investigation into this task. It adopts MPI~\cite{wizadwongsa2021nex} and Plenoctree~\cite{yu2021plenoctrees} representation to represent radiance fields and relies on handcrafting elaborate voxel features to train a foreground-background binary classifier. An important limitation of NVOS is that it demands scene-specific optimization since the binary classifier has to be trained for each scene individually. Thus it cannot generalize to novel scenes. Besides, it requires 3D graph-cut for post-processing, which introduces extra computational overhead during test.  Recently, ISRF~\cite{goel2022interactive} proposes a fast 3D interactive method. It adopts K-means cluster and nearest-neighbor feature matching to obtain high confidence segmented regions of radiance fields and then applies bilateral filter iteratively to obtain final segmentation of radiance fields. An essential step of ISRF is to learn 3D semantic features by distillation from pre-learned 2D semantic features for accurate feature matching, which also demands scene-specific optimization. As a result, ISRF is not capable of generalizing to novel scenes. In this work, we proposes a novel method which makes the first attempt at scene-generalizable interactive segmentation.

\section{Method}
\subsection{Overview}
\vspace{-2pt}
We aim to learn a scene-generalizable interactive segmentation model, which is able to perform accurate object segmentation on unseen scenes represented by radiance fields, guided by the interactive user clicks in a given set of multi-view 2D images.
Our proposed model, referred to as Scene-Generalizable Interactive Segmenter in Radiance Fields (\emph{SGISRF}), takes as input the pre-learned radiance fields $\mathcal{R}$ for a given scene and a series of user clicks $\{C_i\}$ on multi-view 2D images from this scene, and predicts a binary segmentation mask $M$ in 3D space. Following the typical formulation for 2D interactive segmentation~\cite{xu2016deep, sofiiuk2022reviving}, our \emph{SGISRF} models the user interaction as an iterative process and outputs a series of masks $\{M_t\}$ sequentially, thereby refining the segmentation mask iteratively. Formally, \emph{SGISRF} predicts the mask $M_t$ in the $t$-th user interaction based on previous prediction $M_{t-1}$ and user clicks $C_{1:t}$:
\vspace{-4pt}
\begin{equation}
\vspace{-2pt}
    M_t = \mathcal{F}_{\emph{SGISRF}} (\mathcal{R}, M_{t-1}, C_{1:t}),
    \label{eqn:framework}
\end{equation}
where $\mathcal{F}_{\emph{SGISRF}}$ denotes the transformation function of our \emph{SGISRF}. 

As illustrated in Figure~\ref{fig:pineline}, our \emph{SGISRF} encodes sparse user clicks in 2D space into informative 3D guidance features by Cross-Dimension Guidance Propagation, and fuses them with the density features extracted from the radiance fields. The fused features are fed into the Uncertainty-Eliminated 3D segmentation module for efficient yet accurate segmentation. The whole model is optimized effectively with 2D mask annotations under the designed Concealment-Revealed Supervised Learning, which can reveals and correct the concealed 3D segmentation errors resulted from the supervision in 2D space using 2D mask annotations. 

\begin{figure*}[!t]
  \centering
  \includegraphics[width=0.92\textwidth]{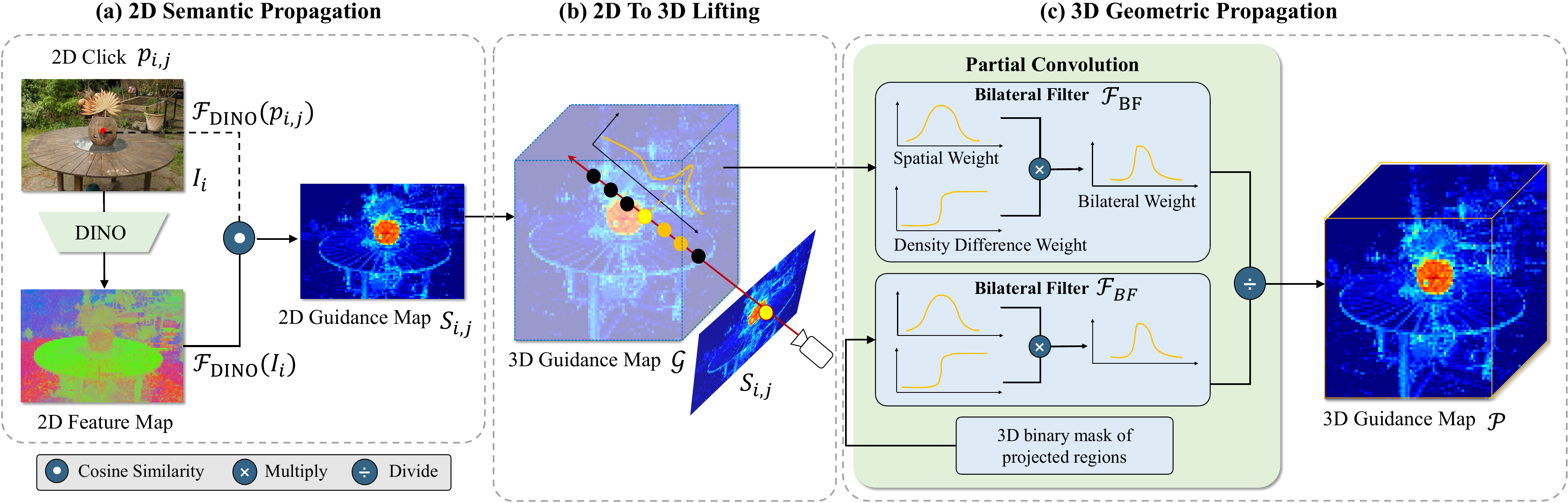}
  \vspace{-12pt}
  \caption{The proposed Cross-Dimension Guidance Propagation first performs 2D propagation leveraging semantic similarities, then lifts the guidance from 2D to 3D. Finally, it perform 3D geometric propagation to obtain informative 3D user guidance.}
\vspace{-10pt}
\end{figure*}
\label{cross dimension propagation fig}

\vspace{-6pt}
\subsection{Volumetric Sampling in Radiance Fields}
\vspace{-2pt}
\label{sec:scene_rep}
Radiance fields can be modeled with a continuous function 
\begin{math}
    (\sigma, \textbf{c} ) = \mathcal{R}(\textbf{x}, \textbf{d}),  
\end{math}
which map a position $\textbf{x} \in \mathbb{R}^3$ in 3D space and a view direction $\textbf{d} \in \mathbb{R}^2$ to a volume density $\sigma \in \mathbb{R}$ and view-depended color $\textbf{c}$ in 3D space. The radiance fields can be used to render 2D images by applying the classical volume rendering formulation~\cite{mildenhall2021nerf}. Formally, the color of a pixel in 2D image $C(\mathbf{r})$ is estimated by sampling $N$ positions along the camera ray $\mathbf{r}$ towards this pixel:
\vspace{-4pt}
\begin{equation}
\vspace{-4pt}
    C(\textbf{r})=\sum_{i=1}^{N} T_i\alpha_i\textbf{c}_i,
    \label{eqn:render}
\end{equation}
where $T_i$ denotes the cumulative transmittance from position $1$ to $i-1$ and $\alpha_i$ is the opacity of position $i$, which are calculated respectively by:
\vspace{-12pt}
\begin{equation}
\vspace{-5pt}
\begin{split}
& T_i = \prod \limits_{i=1}^{n-1}(1-\alpha_i), \quad
 \alpha_{i} = 1 - exp(-\sigma_i\delta_i ).
\label{eqn:opacity}
\end{split}
\end{equation}
Herein, $\delta_i$ is the distance between adjacent positions. 

The continuous function $\mathcal{R}$ can be modeled in various forms such as implicit representation by NeRF~\cite{mildenhall2021nerf} or explicit representations like point cloud~\cite{xu2022point}, voxel grid~\cite{fridovich2022plenoxels} or mesh~\cite{yariv2023bakedsdf}. To extract an effective 3D scene representation from the radiance fields for training our \emph{SGISRF}, following NeSF~\cite{vora2021nesf}, we sample uniform discrete positions in 3D volume and query $\mathcal{R}$ for corresponding volume densities from a pre-learned Radiance Fields to obtain a density grid representation. Note that we transform the density grid to the opacity grid $\mathcal{A}$ by Equation~\ref{eqn:opacity} so that it's normalized to $[0, 1]$. 

\vspace{-6pt}
\subsection{Cross-Dimension Guidance Propagation}
\vspace{-2pt}
\label{sec:propagation}
We aim to perform 3D segmentation relying on only a small number of interactive user clicks, which poses the essential challenge of encoding the scarce 2D user clicks into rich 3D guidance features. To this end, we design the Cross-Dimension Guidance Propagation mechanism, which leverages the feature continuity and similarity in both 2D and 3D spaces as prior knowledge to propagate the user guidance and thus augment the informativeness of the user guidance substantially. It consists of three steps: 2D semantic propagation, 2D to 3D lifting and 3D geometric propagation. 

\noindent\textbf{2D Semantic Propagation.} For a given scene represented by its radiance fields, interactive user clicks (including positive and negative) on multi-view 2D images of the scene are provided to guide the segmentation process iteratively. For each user click, the proposed \emph{SGISRF} performs 2D propagation to estimate a guidance-confidence map over the whole 2D image in the corresponding view. Intuitively, such guidance-confidence map should assign high confidence to those points that are physically similar to the click point while keeping low confidence for dissimilar points. 

Typically the physical similarities between pixels in an image can be well characterized by semantic features. Thus our \emph{SGISRF} performs 2D propagation by measuring the semantic similarities between pixels to obtain the guidance-confidence map. Considering the $j$-th positive click point $p^+_{i,j}$ in a rendering image $\mathcal{I}_i \in \mathbb{R}^{H \times W \times 3}$ from the $i$-th view of a scene, we first construct a semantic feature space $\mathcal{F}_{\text{DINO}}$ by employing the pre-trained DINO model~\cite{caron2021emerging} which is a powerful feature learning model optimized by self-supervised learning. Then we calculate the Cosine similarities between the click point $p^+_{i,j}$ and all pixels in this feature space. leading to the confidence map $S_{i,j}^+ \in \mathbb{R}^{H \times W}$ for the user click $p^+_{i,j}$. Thus the element at $[u,v]$ of $S_{i,j}^+$ is calculated as:
\vspace{-2pt}
\begin{equation}
\vspace{-2pt}
    S_{i,j}^+[u,v] = \text{Cosine} (\mathcal{F}_{\text{DINO}}(p^+_{i,j}), \ \mathcal{F}_{\text{DINO}}(\mathcal{I}_i[u,v])).
\end{equation}
Our \emph{SGISRF} performs 2D propagation by encoding user clicks into a confidence map in a soft manner rather than the hard manner with binary values, thereby preserving rich semantic similarities.

\noindent\textbf{2D to 3D Lifting.} We further lift the obtained confidence map for each user click from 2D to 3D so that we can encode user clicks into 3D guidance representations. Specifically, we project each 2D pixel into 3D volume space and calculate the projected location in line with the volume rendering process shown in Equation~\ref{eqn:render}. Intuitively, the rendered color C(\textbf{r}) for a 2D pixel at $[u, v]$ can be viewed as the weighted sum of the color of sampled points along the camera ray $r$, where the weight for $i$-th point is $T_i\alpha_i$. Hence, we can approximate the 3D projected location for the 2D pixel at $[u, v]$ with the position of sampled point with the maximum weight since this sample contributes most to the rendering of the pixel:
\vspace{-3pt}
\begin{equation}
\vspace{-3pt}
    \mathbf{e} = \text{Coord}(\text{argmax}_{i}(T_i\alpha_i)), \ i=1, \dots, N,
\end{equation}
where $\mathbf{e}\in \mathbb{R}^3$ is the 3D coordinates of the sample with the maximum weight and the function `Coord' returns the 3D coordinates for a sampled point. In practice, we select the projected point $\mathbf{e}$ as well as its 8 nearest neighboring points in the volume grid as the projected 3D area for this 2D pixel considering the spatial continuity.
By constructing the projection between each 2D pixel in a confidence map to 3D volume space, we lift a 2D guidance-confidence map to 3D space by confidence inheritance and obtain a 3D guidance-confidence map.

Each interactive user click produces a 3D guidance-confidence map and there could be overlap between different confidence maps. For each overlapping grid point covered by multiple 3D confidence maps, the max confidence value of all maps is used. Consequently, we fuse all 3D guidance maps of positive and negative user clicks respectively, denoting the fused guidance maps as $\mathcal{G}^+$ and $\mathcal{G}^-$.

\noindent\textbf{3D Geometric Propagation.}  
We further perform 3D propagation by exploiting the geometric continuities. Specifically, we apply the Bilateral Filter~\cite{tomasi1998bilateral} to perform local propagation in both the opacity fields calculated from the density information (shown in Equation~\ref{eqn:opacity}) and the spatial fields. As a result, such propagation takes into account both the geometric and spatial continuities in the neighborhood. Taking the positive guidance propagation as an example, a 3D grid point $x$ can receive the guidance propagated from its neighboring grid points by applying Bilateral Filter $\mathcal{F}_\text{BF}$: 
\vspace{-3pt}
\begin{equation}
\vspace{-6pt}
\begin{split}
&\mathcal{F}_\text{BF}(\mathcal{G}^+, x)=\frac{1}{W}\sum_{x_i\in\Omega}\mathcal{G}^+[x_i]g_\alpha(\left\|\textbf{$\alpha$}_{x_i} - \textbf{$\alpha$}_{x} \right\|) g_s(\left\|x_i - x \right\|),\\
&W = \sum_{x_i\in\Omega}g_\alpha(\left\|\textbf{$\alpha$}_{x_i} - \textbf{$\alpha$}_{x} \right\|) g_s(\left\|x_i - x \right\|),\\
&g_\alpha=exp(-\frac{\left\|\textbf{$\alpha$}_{x_i}-\textbf{$\alpha$}_{x} \right\|^2}{2\sigma^2_\alpha}), \quad
g_s=exp(-\frac{\left\|x_i-x \right\|^2}{2\sigma^2_s}),
\end{split}
\end{equation}
where $\Omega$ is a $3 \times 3 \times 3$ neighboring grids of $x$ and $\alpha_{x_i}$ is the opacity of $x_i$. $g_{\alpha}$ and $g_s$ are the Gaussian smoothing functions for the opacity and spatial fields respectively. $W$ is normalization term. Using Bilateral Filter for field propagation has been previously explored~\cite{goel2022interactive}.

It should be noted that the projected 3D regions from 2D guidance maps only account for a portion of 3D volume, which implies the remaining regions that has no projection from 2D guidance maps have no guidance confidence and thus cannot contribute to the guidance propagation. A straightforward solution is to initialize the guidance values of these un-projected regions as $0$ and apply the Bilateral Filter over the whole 3D space. However, such implementation would result in dilution of propagation and over-smoothness, especially for those points that are surrounded by a number of un-projected points. To address this problem, we propose to normalize the Bilateral Filter by Partial Convolution~\cite{liu2018image}:
\vspace{-5pt}
\begin{equation}
\vspace{-5pt}
\mathcal{P}^+=\frac{\mathcal{F}_\text{BF}(\mathcal{G}^+, x)}{\mathcal{F}_\text{BF}(\mathcal{M}, x) + \epsilon},
\end{equation}
where $M$ is the binary mask indicating the fused projected regions from 2D guidance maps and $\epsilon$ is a small constant value for numerical stability. The negative 3D guidance confidence map $\mathcal{P}^-$ can be obtained in the same way.

\begin{figure}[!t]
  \centering
  \includegraphics[width=0.95\linewidth]{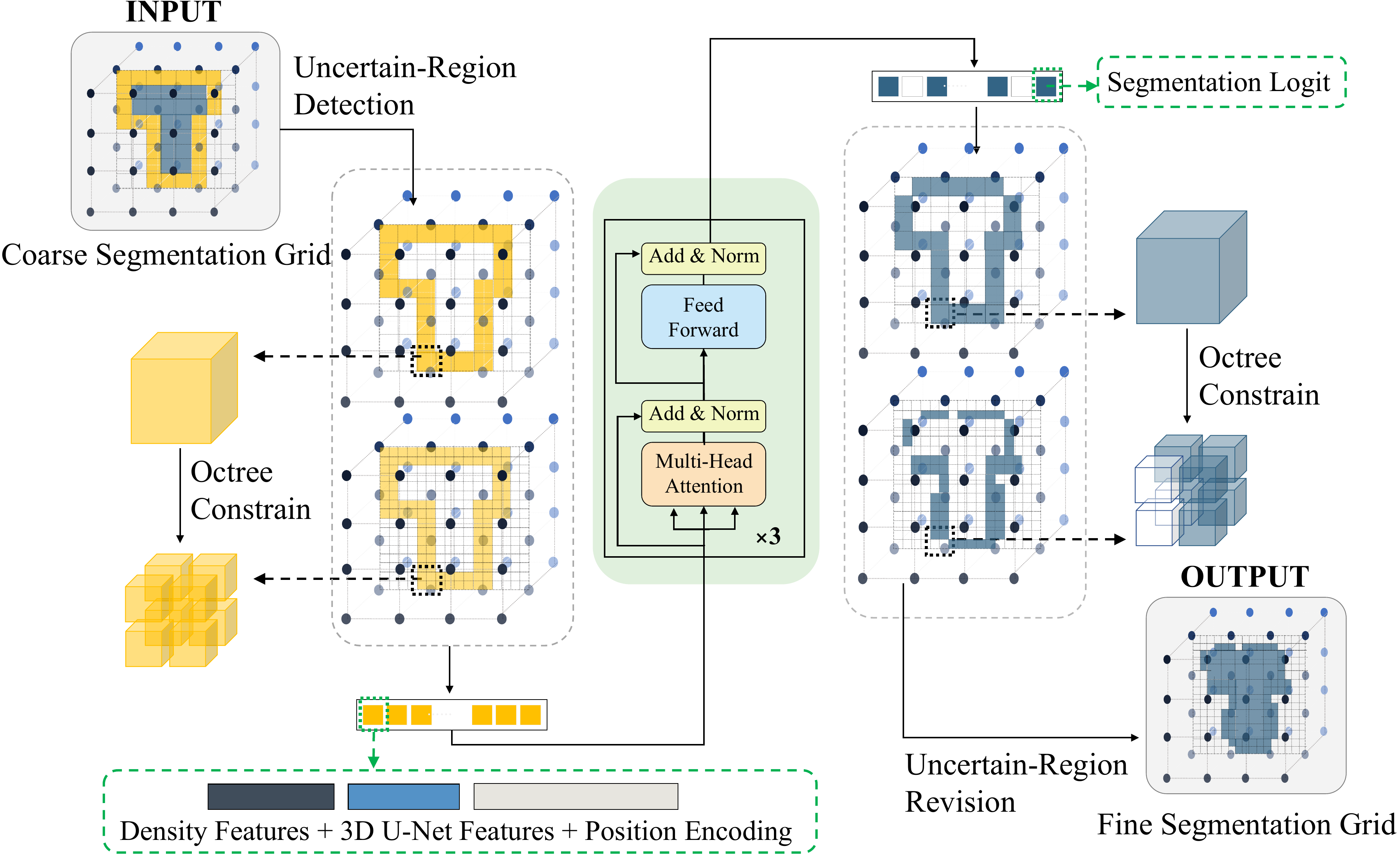}
  \vspace{-12pt}
  \caption{The proposed Uncertainty-Eliminated 3D Segmentation employs Transformer to fuse multi-scale density features for uncertainty refinement.}
  \vspace{-18pt}
  \label{fig:refine}
\end{figure}

\vspace{-5pt}
\subsection{Uncertainty-Eliminated 3D Segmentation}
\vspace{-2pt}
\label{sec:segmentation}
We employ 3D-Unet as our segmentation backbone in 3D space. To be specific, we fuse the 3D scene representations characterized by sampled opacity grid $\mathcal{A}$ (Section~\ref{sec:scene_rep}) and the propagated user guidance $[\mathcal{P^+};\mathcal{P^-}]$ by simple point-wise concatenation. The fused features are fed into 3D-Unet for segmentation, which performs binary classification between foreground and background for each grid point and produces a probability grid of being foreground:
\vspace{-5pt}
\begin{equation}
\vspace{-6pt}
\begin{split}
    \mathcal{M} &= \mathcal{F}_\text{3D-Unet}([\mathcal{A}, \ \mathcal{P^+},  \ \mathcal{P^-}])
    = \text{Sigmoid}(\mathbf{s}),
\end{split}
\label{eqn:unet}
\end{equation}
where $\mathcal{M} \in \mathbb{R}^{X \times Y \times D \times 1}$ is the obtained probability grid, whose values are between $[0,1]$. $X, Y, D$ are the height, width and depth of the sampled 3D grid. $\mathbf{s} \in \mathbb{R}^{X \times Y \times D \times 1}$ denotes the logit values.

Intuitively, performing segmentation in higher-resolution 3D grid space yields more precise performance whilst it incurs heavier computational burden. To perform 3D segmentation accurately yet efficiently, we design Uncertainty-Eliminated 3D Segmentation module, which draws inspiration from Mask Transfiner~\cite{ke2022mask} and adapts it from 2D to 3D. As shown in Figure~\ref{fig:refine}, it first predicts a coarse mask in low-resolution grid space and then refines a fraction of uncertain mask predictions in high-resolution grid space. 
 
\noindent\textbf{Refinement of Uncertainties.} We identify those grid points whose predicted probability of being foreground is close to 0.5 as uncertain predictions. These uncertain regions are mostly near the segmentation boundaries. Our model refines these uncertain predictions by re-predicting their mask labels in higher resolution by Transformer while the remaining certain regions with high confidence directly inherit the predictions from the low-resolution segmentation. As shown in Figure~\ref{fig:refine}, Octree structure is used to build positional correspondence between two resolutions of predictions. In high-resolution 3D segmentation, the scene density features represented by opacity grid $\mathcal{A}$ are sampled in high resolution. Meanwhile, the learned features by 3D-Unet are also leveraged. 

\noindent\textbf{Hierarchical Interaction.}
To facilitate the interaction and knowledge propagation between different resolutions of segmentation, we fuse multi-level scene density features together and fed them into the Transformer, which performs refinement over both low-resolution and high-resolution predictions. We observe that such simple hierarchical interaction can further rectify some false low-resolution predictions. Using Octree, such interaction can be performed quite efficiently with negligible computational overhead.

\vspace{-8pt}
\subsection{Concealment-Revealed Supervised Learning}
\label{sec:supervision}
Due to the lack of 3D segmentation groundtruth, we can only leverage 2D mask annotations to supervise the learning of our \emph{SGISRF}. Thus, we render 3D segmentation results into 2D space to conduct supervision indirectly. We adopt two types of supervision: 1) scene-holistic segmentation supervision, which performs supervision on both foreground and background to push model to distinguish between foreground and background, thereby yielding a basic segmentation; 2) object-exclusive segmentation supervision, which focuses on the segmented foreground by scene-holistic supervision and  eliminates the effect of the background during 3D-to-2D rendering, thus revealing and correcting the segmentation errors concealed during rendering.  

\noindent\textbf{Scene-Holistic Segmentation Supervision.}  
Following the typical rendering way of 3D segmentation in radiance fields~\cite{zhi2021place}, we use the produced segmentation logit values by 3D-Unet in Equation~\ref{eqn:unet} to replace the color values in the volume rendering formulation in Equation~\ref{eqn:render}, which results in the rendering of 2D logit map:
\vspace{-5pt}
\begin{equation}
\vspace{-4pt}
    \hat{s}_{\text{hol}}(\textbf{r})=\sum_{i=1}^{N} T_i\alpha_i s_i,
    \label{eqn:mask-render}
\end{equation}
where $s_i$ is the logit value of the $i$-th sampling position along the camera ray $\mathbf{r}$ in 3D space and  $\hat{s}_\text{hol}(\textbf{r})$ is the rendered logit value for the corresponding pixel in 2D space.  $\hat{s}_\text{hol}(\textbf{r})$ is further used to perform binary classification between foreground and background for 2D segmentation, supervised by Cross-Entropy loss (CE):
\vspace{-3pt}
\begin{equation}
\vspace{-3pt}
\begin{split}
    \mathcal{L}_\text{hol} = &\text{CE}(\hat{s}_{\text{hol}}(\textbf{r}), y)\\
    = & y \log(\text{Sigmoid}(\hat{s}_\text{hol}(\textbf{r}))) \\&+ (1-y)\log(1-\text{Sigmoid}(\hat{s}_\text{hol}(\textbf{r}))),
\end{split}
\end{equation}
where $y$ is the groundtruth of 2D mask for the pixel being rendered.

\begin{figure}[!t]
  \centering
  \includegraphics[width=\linewidth]{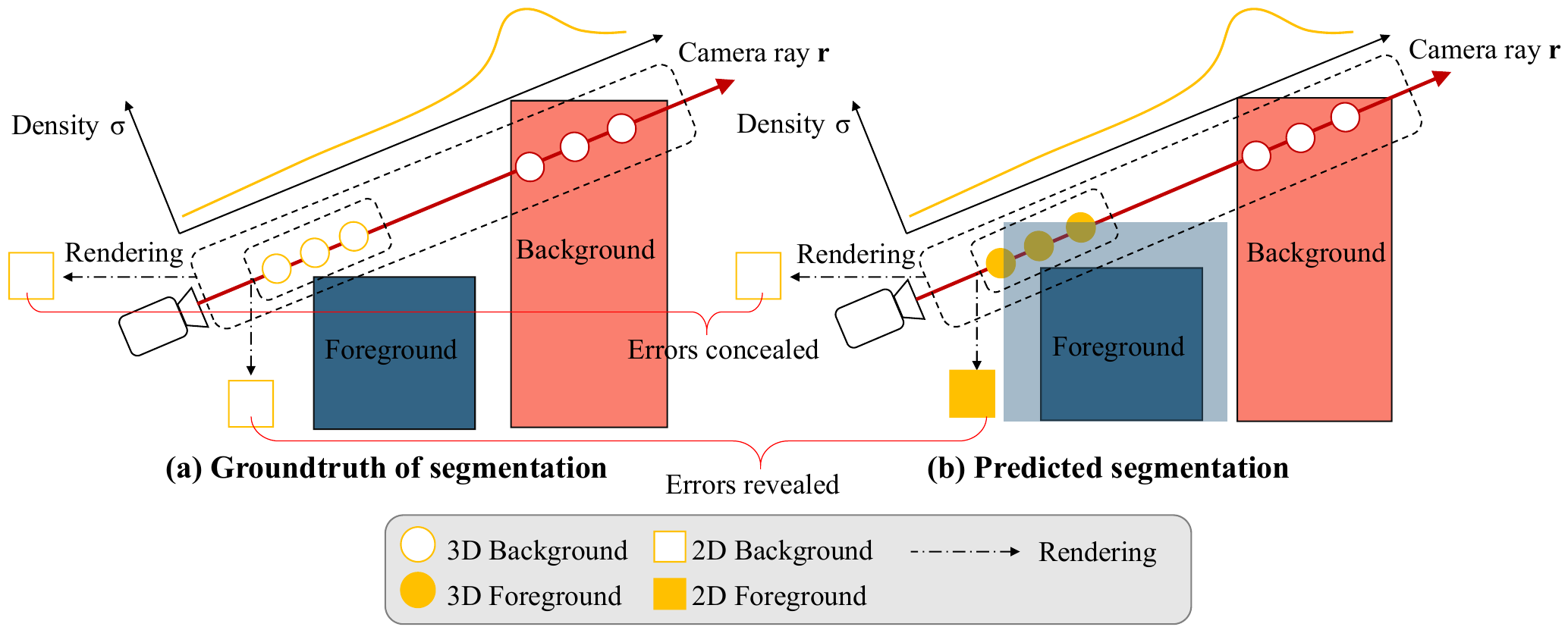}
  \vspace{-22pt}
  \caption{The sampled points in yellow circles in (b) are mis-classified as foreground. However, when rendering from 3D to 2D holistically, the corresponding 2D pixel along the ray $r$ is rendered as background correctly since the other background points (white circles) with larger density along the ray dominate the rendering result. Thus, these segmentation errors of yellow circles are concealed. By re-rendering only the predicted foreground regions into 2D, the mis-classified yellow points along ray $r$ would be rendered into foreground in 2D. Thus, these errors can be revealed and corrected.}
  \vspace{-13pt}
  \label{fig:loss}
\end{figure}

\begin{figure*}[!t]
\centering
\includegraphics[width=0.7\linewidth]{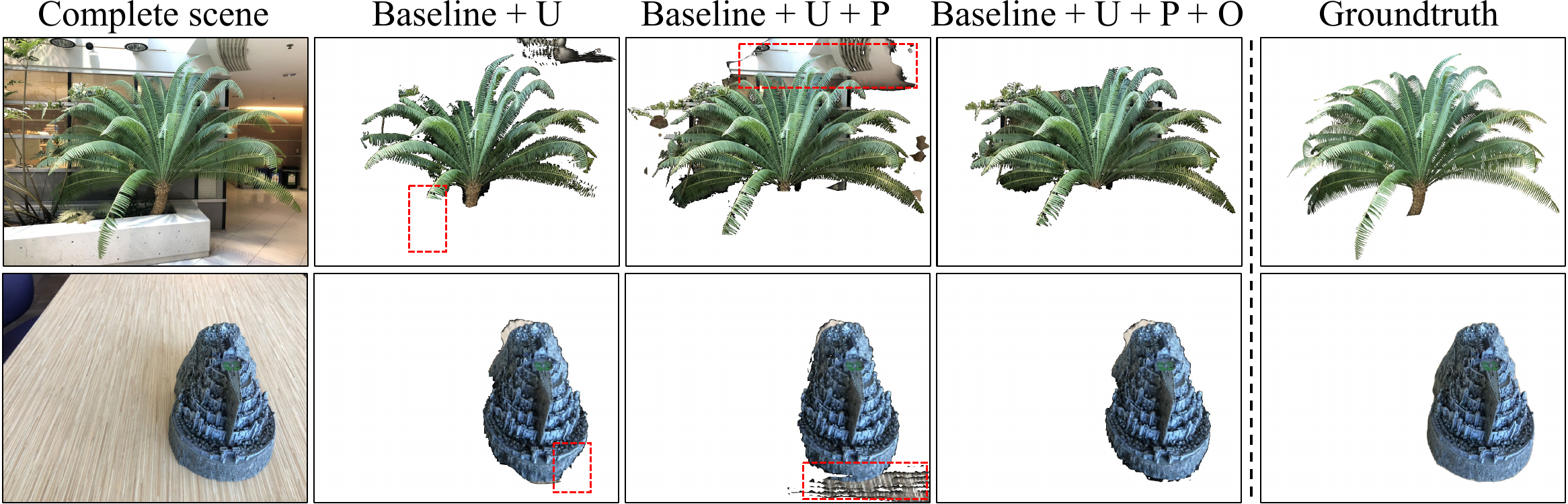}
\vspace{-12pt}
\caption{Visualization of rendering segmented foreground scenes. `U', `P' and `O' denote `Uncertainty-Eliminated 3D segmentation’, `Cross-Dimension Guidance Propagation' and `Object-Exclusive Segmentation Supervision’ respectively. While `Cross-Dimension Guidance Propagation' enables our model to segment more complete foreground, `Object-Exclusive Segmentation Supervision’ can reveal and correct the false positive predictions, which are mostly concealed during 2D supervision.}
\label{fig:ablation1}
\vspace{-8pt}
\end{figure*}

\begin{figure*}[!t]
\centering
\includegraphics[width=0.62\linewidth]{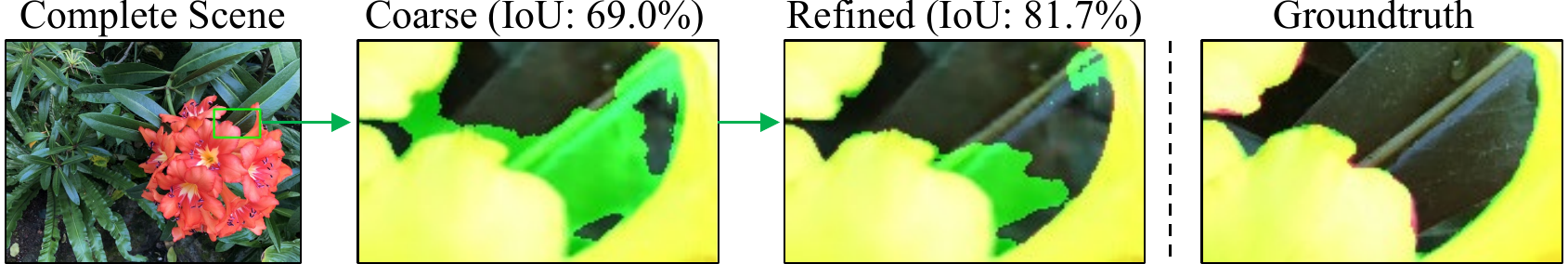}
\vspace{-12pt}
\caption{Effect of the proposed Uncertainty-Eliminated 3D Segmentation.}
\label{fig:ablation2}
\vspace{-12pt}
\end{figure*}

\noindent\textbf{Object-Exclusive Segmentation Supervision.}
Rendering 3D segmentation results into 2D compresses the 3D segmentation logit values into 2D by aggregating the 3D sampled points along each camera ray into a single 2D pixel as shown in Equation~\ref{eqn:mask-render}. As a result, substantial detailed 3D segmentation information is lost during rendering, which may conceal some prediction errors of 3D segmentation and thereby fails to supervise these error during optimization. Consider the example shown in Figure~\ref{fig:loss} (b), the sampled background points (yellow circles) along the camera ray $r$ are mis-classified as foreground. however, the corresponding 2D pixel are rendered as background correctly (white square) since the other background points (white circles) with larger density along the ray dominate the rendering result. Consequently, these 3D segmentation errors for yellow points are concealed and thus no supervision is conducted on them for correction.

To deal with above limitation, we propose the object-exclusive segmentation supervision, which re-renders the 3D segmentation results only within the predicted foreground regions under the scene-holistic supervision and conducts supervision on the obtained 2D mask. Thus, the white background points are not considered during rendering and the mis-predicted foreground 3D points in Figure~\ref{fig:loss} (b) would be rendered as foreground (yellow square) in 2D, which are inconsistent with the groundtruth. Thus these prediction errors can be revealed and corrected by such supervision.

To be specific, we design two losses for object-exclusive supervision. First, we directly apply the rendering method based on segmentation logit values shown in Equation~\ref{eqn:mask-render} whilst only the predicted foreground regions are rendered:
\vspace{-5pt}
\begin{equation}
\vspace{-5pt}
    \hat{s}_{\text{obj}}(\textbf{r})=\sum_{i=1}^{N} \mathcal{M}^{\text{hol}}_i \cdot T_i\alpha_i s_i,
    \label{eqn:fore-render}
\end{equation}
where $\mathcal{M}^\text{hol}$ is the predicted mask under the scene-holistic supervision. Note that we use the soft values of $\mathcal{M}$ instead of quantized binary mask values to ease the gradient propagation. The supervision is conducted via Cross-Entropy loss:
\vspace{-5pt}
\begin{equation}
\vspace{-5pt}
    \mathcal{L}_\text{obj}^1 =\text{CE}(\hat{s}_{\text{obj}}(\textbf{r}), y).
\end{equation}
Such supervision propagates gradients to both the segmentation logit $\mathbf{s}_i$ and the prediction mask $\mathcal{M}$. To ease the convergence of optimization, we also design another loss for object-exclusive supervision which drops the segmentation logit and only uses the opacity for rendering, resulting in 2D occupied field $\hat{o}$:
\vspace{-5pt}
\begin{equation}
\vspace{-5pt}
    \hat{o}(\textbf{r})=\sum_{i=1}^{N} \mathcal{M}^{\text{hol}}_i \cdot T_i\alpha_i.
    \label{eqn:mask-render2}
\end{equation}
$L_2$ loss is applied to conduct supervision:
\vspace{-4pt}
\begin{equation}
\vspace{-4pt}
    \mathcal{L}_{obj}^2=\left\| \hat{o}(\textbf{r}) - y \right\|^2.
\end{equation}

Combining the scene-holistic supervision and object-exclusive supervision together, our \emph{SGISRF} is optimized by:
\vspace{-4pt}
\begin{equation}
\vspace{-4pt}
    \mathcal{L} = \lambda\mathcal{L}_\text{hol} + \lambda_{1}\mathcal{L}_\text{obj}^1 + \lambda_{2}\mathcal{L}_\text{obj}^2.
\end{equation}

\section{EXPERIMENTS}
\subsection{Datasets and Exeperimental Setup}
\noindent\textbf{Datasets.} \emph{LLFF}~\cite{mildenhall2019local} provides a front-facing real-world scene dataset named Real Iconic, which contains 47 real-world scenes. Following NVOS~\cite{ren2022neural}, we use \emph{NeRF-LLFF}, a subset of Real Iconic including 8 scenes, as the test set. To perform scene-generalizable evaluation, we use the remaining 39 scenes in Real Iconic as the training data. Note that NVOS performs both training and test on the same set of 8 scenes included in \emph{NeRF-LLFF}. Since the mask annotations are not provided in Real Iconic, we annotate masks manually using PeddleSeg~\cite{liu2021paddleseg}. 
\emph{CO3D}~\cite{reizenstein2021common} is a $360^{\circ}$-view real-world dataset containing 19k objects of 50 categories, each scene containing approximately 200 images.
Considering the experimental efficiency, we randomly select 180 scenes from the dataset to construct a small dataset for performing our experiments, among which 150 scenes are used as training set and the remaining 30 scenes as test set.

\begin{figure*}[htb]
\centering
\includegraphics[width=0.72\linewidth]{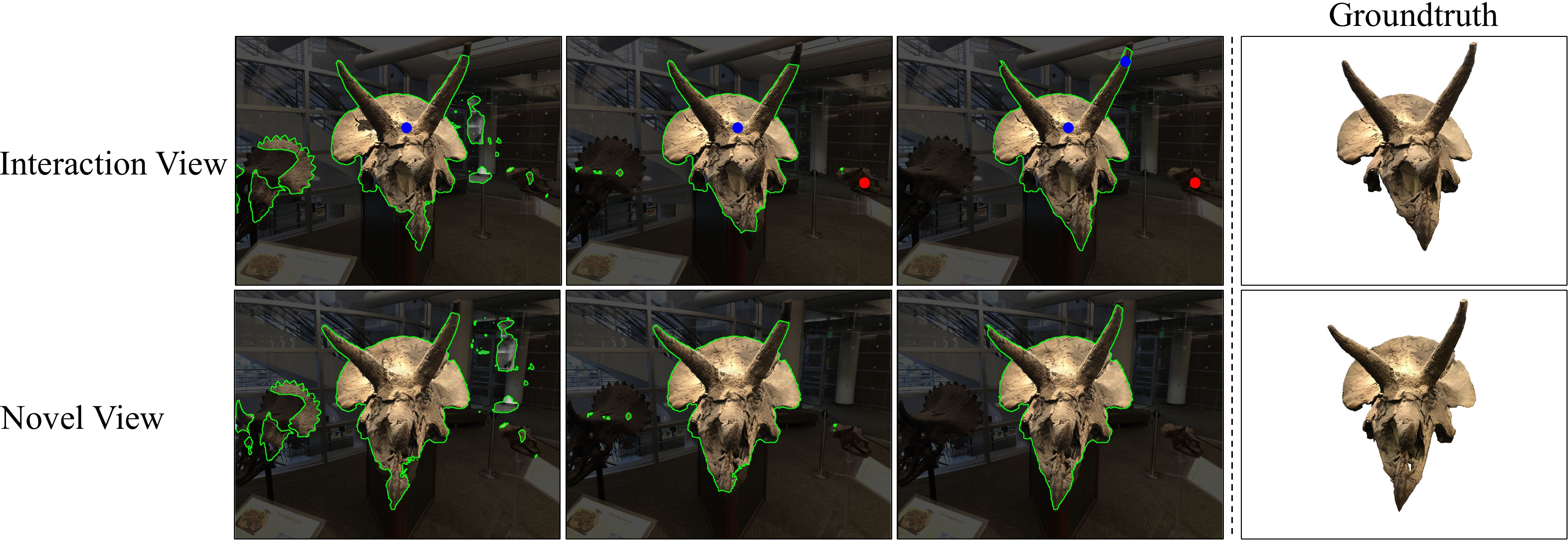}
\vspace{-8pt}
\caption{Progressive Segmentation by our \emph{SGISRF} with positive (blue points) and negative (red points) user clicks.}
\label{fig:click_vis}
\vspace{-10pt}
 \end{figure*}

\noindent\textbf{ Implementation Details.} We combine the training sets of Real Iconic and CO3D together to train our model. Following NeSF~\cite{vora2021nesf}, we only perform user interaction on a few of views for each scene. Specifically, user interactions are conducted on 20 and 5 views for CO3D and Real Iconic, respectively.
In training stage and ablatation study, we simulate the interactive clicks by programs following ~\cite{sofiiuk2022reviving}, in comparison between our model and other methods , we perform human interaction to explore the performance limit of our model. This accounts for the performance difference of \emph{NeRF-LLFF} between Table \ref{tab:ablation} and Table \ref{tab:comparison}.
We train our model for 25k iterations using Adam~\cite{kingma2014adam} with initial learning rate of $10^{-3}$. The weights of the losses are tuned to be $\lambda=1.0$, $\lambda_{1}=1.0$ and $\lambda_{2}=1.0$.
We re-implement NVOS since the source code is not released.

\noindent\textbf{ Evaluation Criteria.} Since the 3D mask annotations are not available for evaluation, following NVOS~\cite{ren2022neural}, we render the 3D segmentation results into 2D and adopt two 2D metrics for evaluation: pixel-wise classification accuracy (\emph{Acc}) and foreground intersection-over-union (\emph{IoU}). Besides, we also evaluate 3D segmentation by rendering the radiance fields of the segmented 3D foreground into 2D and evaluating the rendered image quality. We use SSIM~\cite{wang2004image}, PSNR, and LPIPS~\cite{zhang2018unreasonable} as evaluation metrics.

\vspace{-5pt}
\subsection{Ablation Study}
\label{sec:ablation}
\vspace{-2pt}

We first conduct ablation studies to investigate the effectiveness of each core design of our \emph{SGISRF}. Abbreviating `Cross-Dimension Guidance Propagation', `Uncertainty-Eliminated 3D Segmentation' and `Object-Exclusive Segmentation Supervision' as `P', `U' and `O' respectively, we ablate all these modules from \emph{SGISRF} to obtain `Baseline' variant for comparison. Note that it is optimized only by the Scene-Holistic Segmentation Supervision. Then we augment the `Baseline' by equipping it with three modules incrementally and measure the performance gain yielded by each module.
Table~\ref{tab:ablation} shows the quantitative comparison while qualitative evaluation on these modules is presented in Figure~\ref{fig:ablation1} and~\ref{fig:ablation2}.

\noindent\textbf{Effect of Cross-Dimension Guidance Propagation.} The performance gap between `Baseline+U' and `Baseline+U+P' in Table~\ref{tab:ablation}, namely $2.1\%$ in Acc and $3.7\%$ in IoU, manifests the effectiveness of Cross-Dimension Guidance Propagation. Besides, the visual comparison between `Baseline+U' and `Baseline+U+P' in Figure~\ref{fig:ablation1} shows that the proposed guidance propagation scheme enables our model to segment more complete foreground, e.g., the highlighted regions in the red dashed boxes. On the other hand, such propagation may lead to over-propagation in some cases, in which some background area near the boundaries could be mis-classified as foreground.  Nevertheless, most of these cases can be well addressed by the proposed Object-Exclusive Segmentation Supervision.

\begin{table}[!t]
  \caption{Abation study on NeRF-LLFF. `U', `P' and `O' denote `Uncertainty-Eliminated 3D segmentation’, `Cross-Dimension Guidance Propagation' and `Object-Exclusive Segmentation Supervision’, respectively.}
\vspace{-12pt}
  \begin{tabular}{l|cc}
    \toprule
Variants \ & Acc$\uparrow$         & IoU$\uparrow$    \\ \midrule
Baseline   &90.3 & 65.0  \\ \midrule
Baseline + U & 91.1    &  65.5    \\
Baseline + U + P  & 93.2    &  69.2    \\
Baseline + U + P + O (Intact \emph{SGISRF}) & \textbf{97.2}    &  \textbf{84.5} \\ 
  \bottomrule
\end{tabular}
\label{tab:ablation}
\end{table}

\begin{figure}[!t]
\centering
\includegraphics[width=0.70\linewidth]{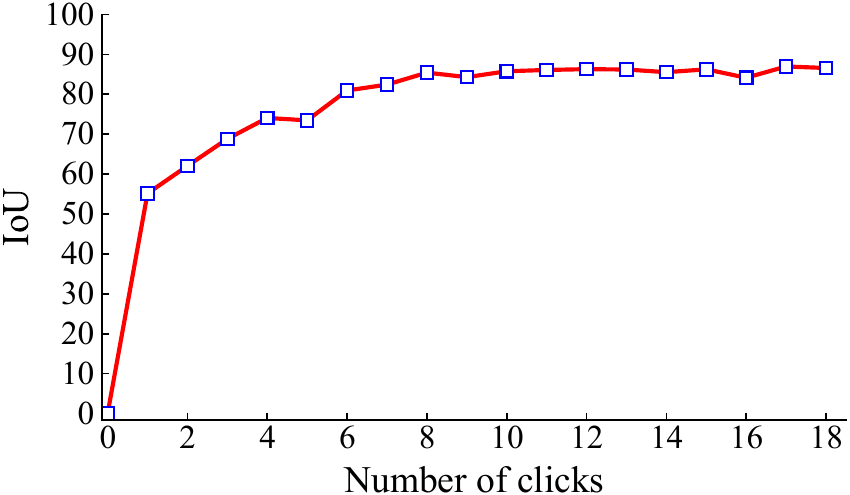}
\vspace{-10pt}
\caption{Foreground IoU by our \emph{SGISRF} for varying number of clicks on NeRF-LLFF.}
\label{fig:curve}
\vspace{-16pt}
\end{figure}

\noindent\textbf{Effect of Object-Exclusive Segmentation Supervision.} Comparing between `Baseline+U+P' and the intact \emph{SGISRF} model in Table~\ref{tab:ablation}, the proposed Object-Exclusive Supervision boosts the performance significantly, especially in `IoU'. The large improvement mainly results from the revealing and correcting of those concealed 3D segmentation errors caused by the information loss during 2D supervision. As discussed before, another interesting observation is that Object-Exclusive Supervision can substantially mitigate the adverse effect of the proposed guidance propagation, namely the potential over-propagation, as illustrated in Figure~\ref{fig:ablation1}.

\begin{figure*}[!t]
\centering
\includegraphics[width=0.83\linewidth]{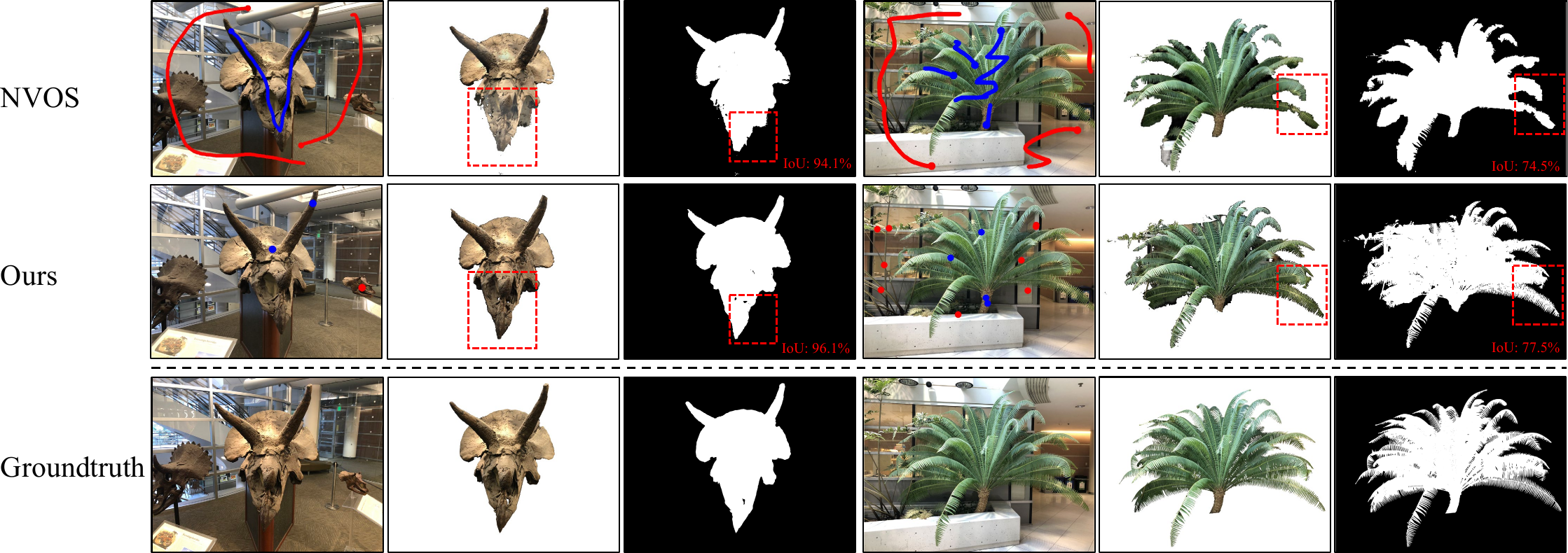}
\vspace{-11pt}
\caption{Visual Comparison between our \emph{SGISRF} and NVOS on NeRF-LLFF. Both rendering of segmented foreground and the 2D mask are presented. The differences are highlighted by red dashed boxes. The positive and negative user interactions are indicated by blue and red points/strokes, respectively.}
\label{fig:vis-llff}
\vspace{-9pt}
\end{figure*}

\begin{figure*}[!t]
\centering
\includegraphics[width=0.79\linewidth]{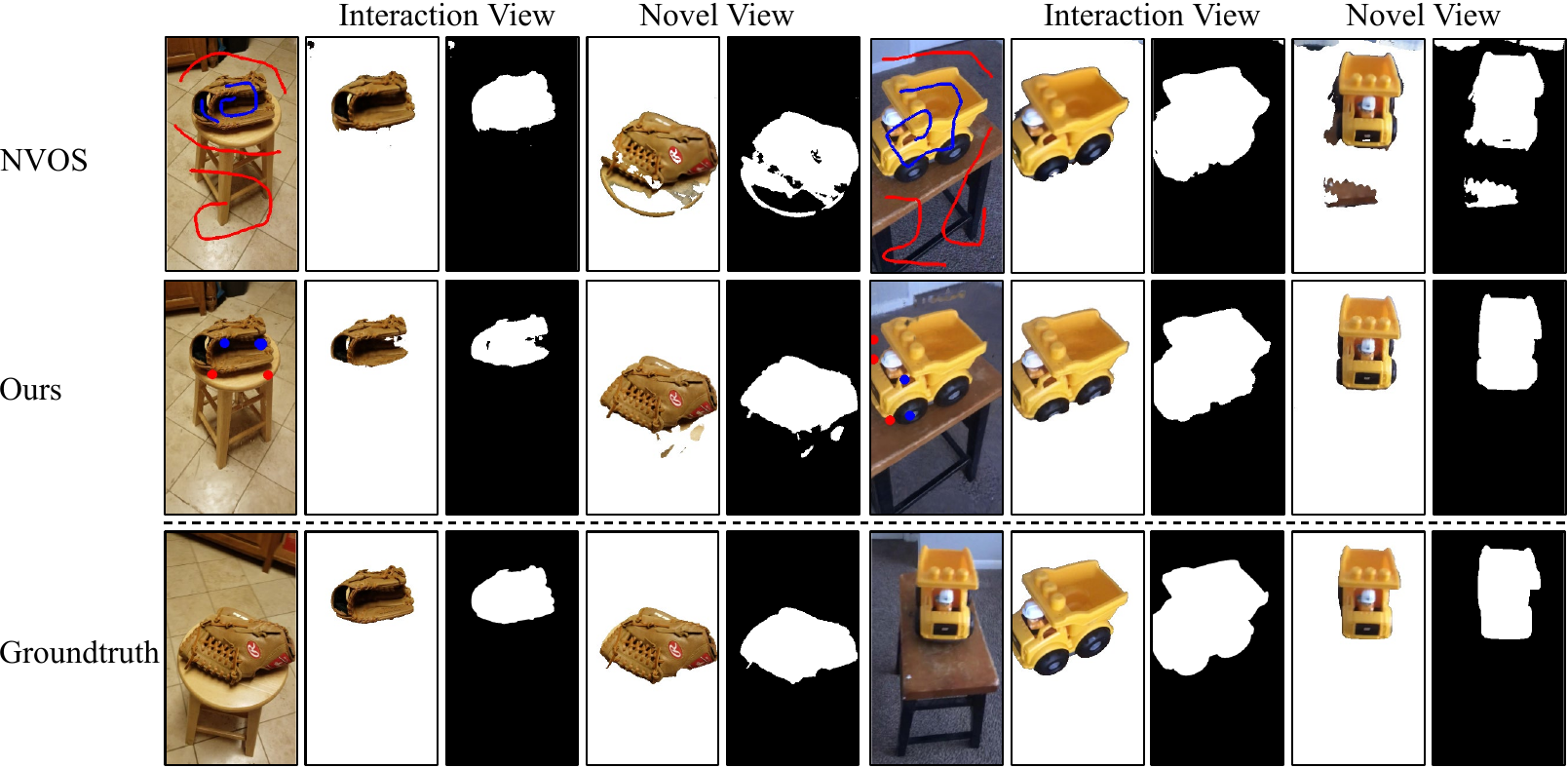}
\vspace{-8pt}
\caption{Visual Comparison between our \emph{SGISRF} and NVOS on CO3D dataset. Our model shows much stronger robustness than NVOS on scenes with large view variation between the interaction view and the novel view.}
\label{fig:vis-co3d}
\vspace{-10pt}
\end{figure*}

\noindent\textbf{Effect of Uncertainty-Eliminated 3D Segmentation.} Table~\ref{tab:ablation} shows the Uncertainty-Eliminated 3D segmentation module yields relatively smaller performance gain than the other two modules primarily due to two reasons. First, this module is particularly effective on highly challenging segmentation scenarios involving complicated boundaries as shown in Figure~\ref{fig:ablation2}, which accounts for only a small portion of cases. Second, the lack of 3D mask annotations for 3D supervision limits its potential to some degree.  

\noindent\textbf{Effect of Interactive User Guidance.} Benefiting from the effective design of our \emph{SGISRF}, especially the proposed Cross-Dimension Guidance Propagation and Object-Exclusive Segmentation Supervision, our \emph{SGISRF} exhibits highly efficient user interactive performance. As shown in Figure~\ref{fig:curve}, \emph{SGISRF} reaches the saturation performance rapidly with only a few user clicks. The example in Figure~\ref{fig:click_vis} also qualitatively demonstrates the highly efficient interactive performance of our \emph{SGISRF}.

\begin{table}
  \caption{Comparison between our \emph{SGISRF} and NVOS on NeRF-LLFF and CO3D datasets.}
  \vspace{-12pt}
  \begin{tabular}{l|cc|cc}
    \toprule
\multirow{2}{*}{Methods} & \multicolumn{2}{c|}{NeRF-LLFF} &\multicolumn{2}{c}{CO3D} 
\\
 & Acc$\uparrow$         & IoU$\uparrow$ 
& Acc$\uparrow$         & IoU$\uparrow$\\ 
\midrule
NVOS    &  92.0            &  70.1   & 95.0 & 79.2 \\
\midrule
Ours    &  \textbf{97.6}        &   \textbf{86.4}   &  \textbf{97.5}  & \textbf{85.8} \\
  \bottomrule
\end{tabular}
\label{tab:comparison}
\vspace{-12pt}
\end{table}

\begin{table}
  \caption{Comparison of rendered image quality of segmented foreground scenes by our \emph{SGISRF} and NVOS on NeRF-LLFF.}
  \vspace{-12pt}
  \begin{tabular}{l|ccc}
    \toprule
Metrics & SSIM$\uparrow$         & PSNR$\uparrow$  & LPIPS$\downarrow$\\ 
\midrule
NVOS    &  0.767            &  18.40   & \textbf{0.213}   \\
\midrule
Ours    &  \textbf{0.768}            &  \textbf{19.55}    & 0.227       \\
  \bottomrule
\end{tabular}
\label{tab:comparison2}
\vspace{-12pt}
\end{table}



\vspace{-5pt}
\subsection{Comparison with NVOS.} 
\vspace{-2pt}
Considering that our \emph{SGISRF} is the first method for interactive segmentation of radiance fields that can be generalizable across different scenes, we compare it with NVOS~\cite{ren2022neural}, a prominent method  demanding scene-specific optimization. NVOS has seen all test scenes during training. In contrast, all test scenes are novel for our \emph{SGISRF}. Thus, it is not strictly a fair comparison. 
Besides, our method only needs a few user clicks for guidance while NVOS requires sufficient user scribbles to learn a binary classifier. Comparison with more methods can be found in our appendix.

\noindent\textbf{Quantitative Comparison.} As shown in Table~\ref{tab:comparison}, our \emph{SGISRF} outperforms NVOS substantially on both datasets in terms of both Acc and IoU, which demonstrates the superiority of our model over NVOS despite unfair comparative setting to our model. Besides, we also compare the rendered image quality of segmented foreground scenes by two methods in Table~\ref{tab:comparison2}. Our \emph{SGISRF} performs on par with NVOS considering the overall performance on three metrics. 

\noindent\textbf{Qualitative Comparison.} We also conduct qualitative comparison between our \emph{SGISRF} and NVOS on both NeRF-LLFF and CO3D in Figure~\ref{fig:vis-llff} and~\ref{fig:vis-co3d}, respectively. These comparisons clearly show the advantages of our model over NVOS. In particular, we observe that our model shows much stronger robustness than NVOS on CO3D scenes with large view variation between the interaction view and the novel view. As shown in Figure~\ref{fig:vis-co3d}, both models perform well on the interaction view. However, the performance of NVOS degenerates distinctly on the novel view in both cases while our model performs consistently well on the novel view.

\section{Conclusion}
In this work we have made the first attempt at scene-generalizable interactive segmentation of radiance fields and propose a novel method, namely \emph{SGISRF}, which can perform 3D scene segmentation on novel scenes without scene-specific optimization. In particular, we propose three novel designs to address three crucial challenges including effective user guidance propagation, efficient 3D segmentation modeling and specially designed 2D supervision scheme. Extensive experiments on two challenging real-world datasets have validated the effectiveness of the proposed \emph{SGISRF}.


\vspace{-3pt}
\bibliographystyle{ACM-Reference-Format}
\bibliography{sample-sigconf}

\appendix
\section{Implementation Details}
\noindent\textbf{Iterative training strategy.}
Inspired by the iterative training strategy in 2D interactive image segmentation~\cite{sofiiuk2022reviving}, we generate user clicks online by programs to simulate human interactive behavior. During each training iteration, our model randomly samples a view of the sampled scene as the interactive view, and then generates a user click based on the groundtruth object mask.Then our model predicts 3D segmentation and renders the predicted mask into 2D space to perform supervision. Such process is iterated for T times to simulate T interactive user clicks. Note that the gradient back-propagation is only performed at the T-th step to improve the optimization efficiency.

\section{Additional Ablation Study}
\begin{figure*}[!t]
\centering
\includegraphics[width=0.6\linewidth]{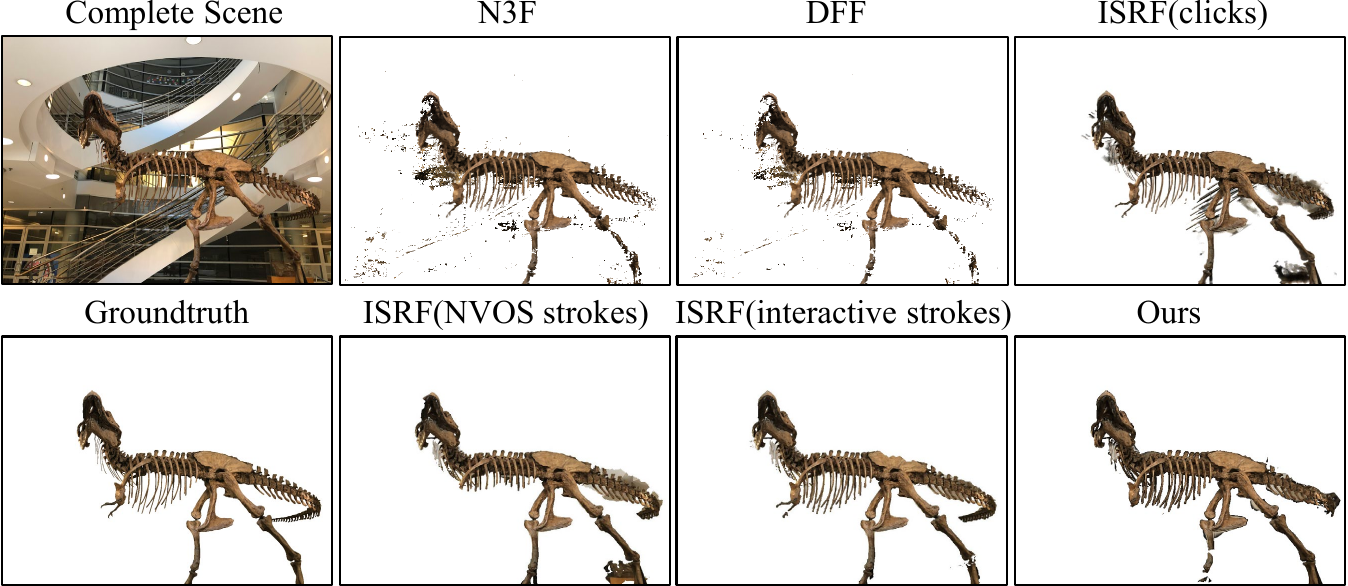}
\vspace{-10pt}
\caption{Visual Comparison with N3F, DFF and ISRF.}
\label{fig:appedix-comparsion-llff}
\vspace{-10pt}
\end{figure*}
\begin{table}[!t]
  \caption{Ablation Study of two losses ($\mathcal{L}_\text{obj}^1$+$\mathcal{L}_\text{obj}^2$) of the proposed Object-Exclusive Segmentation Supervision on NeRF-LLFF. `U', `P'  and `O' denote `Uncertainty-Eliminated 3D segmentation', `Cross-Dimension Guidance Propagation' , `Object-Exclusive Segmentation Supervision', respectively.}
\vspace{-12pt}
  \begin{tabular}{l|cc}
    \toprule
Variants \ & Acc$\uparrow$         & IoU$\uparrow$    \\ \midrule
Baseline + U + P   &93.2 & 69.2  \\ \midrule
Baseline + U + P + $\mathcal{L}_\text{obj}^1$  & 97.0 & 82.5  \\
Baseline + U + P + $\mathcal{L}_\text{obj}^2$  & 96.1 & 78.9 \\ \midrule
Baseline + U + P + O ($\mathcal{L}_\text{obj}^1$+$\mathcal{L}_\text{obj}^2$) & \textbf{97.2}    &  \textbf{84.5} \\ 
  \bottomrule
\end{tabular}
\label{tab:ablation addition object loss}
\vspace{-16pt}
\end{table}
\begin{table}
  \caption{Ablation Study on Dependence of 2D
Semantic Features.}
  \vspace{-12pt}
  \begin{tabular}{l|cc|cc}
    \toprule
\multirow{2}{*}{Features} & \multicolumn{2}{c|}{Ours} &\multicolumn{2}{c}{ISRF} 
\\
 & Acc$\uparrow$         & IoU$\uparrow$ 
& Acc$\uparrow$         & IoU$\uparrow$\\ 
\midrule
DINO    &  \textbf{97.6}            &  \textbf{86.4}   & \textbf{98.2} & \textbf{90.8} \\
LSeg    &  92.5        &   69.5   &  81.0  & 52.7 \\
ResNet-50    & 96.2        &   78.6   &  95.7  & 79.4 \\
  \bottomrule
\end{tabular}
\label{tab:ablation addition features}
\vspace{-16pt}
\end{table}
\begin{table}[!t]
  \caption{Comparison with more methods on NeRF-LLFF.}
\vspace{-12pt}
  \begin{tabular}{l|cc}
    \toprule
Methods \ & Acc$\uparrow$         & IoU$\uparrow$    \\ \midrule
N3F~\cite{tschernezki2022neural}  & 89.2 & 63.3  \\
DFF~\cite{kobayashi2022decomposing}   &90.4 & 66.2  \\
ISRF(clicks)  & 96.7 & 82.7  \\
ISRF(NVOS strokes)  & 96.4 & 83.8 \\ 
ISRF(interactive strokes) & \textbf{98.2}    &  \textbf{90.8} \\ \midrule
Ours & 97.6    &  86.4 \\ 
  \bottomrule
\end{tabular}
\label{tab:comparision addition more methods}
\vspace{-16pt}
\end{table}
\begin{table}[!t]
  \caption{Comparison of time performance with other methods. Our method doesn't require scene-specific optimization.}
\vspace{-12pt}
  \begin{tabular}{l|cc}
    \toprule
Methods \ & \makecell[c]{Time for \\ scene-specific optimization}         & \makecell[c]{Inference time for \\ one interactive step}    \\ \midrule
NVOS   & $\approx$4 mins & 110 secs in total  \\
ISRF  &$\approx$2.5 mins & 3.3 secs  \\
SPIn-NeRF~\cite{mirzaei2023spin}  &$\approx$3-6 mins & 1 secs \\ \midrule
Ours & 0    &  2.5 secs \\ 
  \bottomrule
\end{tabular}
\label{tab:comparision addition time}
\vspace{-20pt}
\end{table}
\begin{table}[!t]
  \caption{Evaluation on More Datasets.}
\vspace{-12pt}
  \begin{tabular}{l|cc}
    \toprule
Datasets \ & Acc$\uparrow$         & IoU$\uparrow$    \\ \midrule
SPIn-NeRF  & 98.7 & 81.3  \\
Tanks and Temples   &91.0 & 68.2  \\
Mip-NeRF 360  & 89.8 & 40.8  \\
  \bottomrule
\end{tabular}
\label{tab:evaluation on more datasets}
\vspace{-16pt}
\end{table}

\subsection{Ablation Study on Two Losses of Object-Exclusive Segmentation Supervision}
The proposed Object-Exclusive Segmentation Supervision consists of two losses:
\begin{equation}
\begin{split}
&\mathcal{L}_\text{obj}^1 =\text{CE}(\hat{s}_{\text{obj}}(\textbf{r}), y), \\
&\mathcal{L}_\text{obj}^2=\left\| \hat{o}(\textbf{r}) - y \right\|^2,\\
\end{split}
\end{equation}
Two losses correspond to different rendering methods. As shown in Equation 11 in the paper, for the first loss $\mathcal{L}_\text{obj}^1$, the segmentation logit values are used for rendering the 2D mask. In contrast, only the opacity rather than the segmentation logit is used for rendering for the second loss $\mathcal{L}_\text{obj}^2$, which results in the 2D occupied field, as shown in Equation~13. To investigate the effect of each loss, we conduct additional ablation study, which is presented in Table~\ref{tab:ablation addition object loss}. 

The results show that both losses boost the performance significantly while $\mathcal{L}_\text{obj}^1$ yields relatively more performance gain than $\mathcal{L}_\text{obj}^2$. This is reasonable since both losses follows the same rationale: only rendering the predicted foreground regions to eliminate the effect of the background regions during the 3D-to-2D rendering, thus revealing the potentially concealed errors.

\subsection{Ablation Study on Dependence of 2D Semantic Features}
 We investigate the effect of three classical semantic features on 2D propagation of our model on NeRF-LLFF dataset, including DINO~\cite{caron2021emerging}, LSeg~\cite{li2022language} and ResNet-50~\cite{long2015fully}. Besides, we also test the effect of these features on ISRF. Table \ref{tab:ablation addition features} shows that: 1) both our model and ISRF perform best on DINO, presumably because of the effectiveness of DINO. 2) ISRF performs worse than our model with LSeg feature, which implies that ISRF is more sensitive to (depends heavier on) the semantic features than our model.

\section{Comparison with more methods}
We emphasize that our model makes the first attempt for scene-generalizable interactive radiance segmentation (all test scenes are unseen during training) whilst all existing methods for interactive radiance segmentation, including NVOS and ISRF, require scene-specific optimization (all test scenes are used for optimization). Thus, it is not fair to compare our method to these methods directly.

Despite the unfair comparison, we compare our method with ISRF on NeRF-LLFF dataset.  Besides, we also compare our method with other radiance fields segmentaion methods~\cite{kobayashi2022decomposing, tschernezki2022neural} by image patch query. The results are shown in Table \ref{tab:comparision addition more methods} and  Figure \ref{fig:appedix-comparsion-llff}. Table \ref{tab:comparision addition more methods} shows that: 1) our model outperforms ISRF distinctly when both methods utilize user clicks as interactive guidance. 2) our model also performs better than ISRF when it is guided by the user strokes utilized by NVOS. 3) ISRF performs better than our model when it carefully selects the user strokes interactively. These encouraging results reveal the effectiveness and potential of our model.

\section{Evaluation on more datasets}
We further evaluate our pre-trained model on three datasets \emph{SPIn-NeRF} ~\cite{mirzaei2023spin}, \emph{Tanks and Temples}~\cite{knapitsch2017tanks} and \emph{Mip-NeRF 360}~\cite{barron2022mip} without any fine-tuning in Table \ref{tab:evaluation on more datasets}. Note that we only test four scenes (Family, Truck, Ignatius and Horse) with relatively good reconstruction quality of \emph{Tanks and Temples}, we use the ground truth masks provided by ~\cite{mirzaei2023spin} for \emph{SPIn-NeRF} while annotate masks for \emph{Tanks and Temples} and \emph{Mip-NeRF 360} by ourselves due to lack of public masks. The performance of \emph{SPIn-NeRF} is actually quite encouraging considering semantic and distribution gap between the training and test data. Our pre-trained model shows poor performance on some scenes of \emph{Tanks and Temples} and \emph{Mip-NeRF 360} due to our training data lacks of complex indoor scenes and large scale outdoor scenes.

\section{Time performance}
We compare the efficiency of different methods in Table \ref{tab:comparision addition time}. Note that all other methods require scene-specific optimization, which generally demands a lot more time than that for inference. NVOS performs segmentation in one time instead of interactive segmentation. Table \ref{tab:comparision addition time} reveals that our method is much more efficient than other methods, benefitting from the scene-generalizable capability.

\section{Limitation}
Our model has two potential limitations: 1) Our model is designed to have scene-generalizable capability. Intuitively, using more training data with more diverse scenes leads to better generalibility of our model. Thus, effective optimization of our model requires large amount of multi-view segmentation data for training. However, due to the limited size of publicly available data for training, the potential of our model has not been fully explored yet. 2) Our model performs 2D propagation in the semantic feature space constructed by DINO. For those scenes with similar DINO features between foreground and background, our model tends to propagate falsely. In these cases, we need to investigate more effective 2D propagation schemes.

\end{document}